\documentclass[sigconf]{acmart}

\usepackage[tight]{subfigure}
\usepackage{xspace}
\usepackage{latexsym}
\usepackage{booktabs}
\usepackage{enumitem}
\usepackage{graphicx,subfigure}
\usepackage{amsmath}
\usepackage{multirow}
\usepackage{amssymb}

\usepackage[ruled,vlined, linesnumbered]{algorithm2e}
\usepackage{epstopdf}
\usepackage{array}
\usepackage{url}
\usepackage{CJKutf8}
\usepackage{color,xcolor}

\newcommand{\eat}[1]{}
\newcommand{\paratitle}[1]{\vspace{1.5ex}\noindent \textbf{#1}}
\newcommand{\ie}{\emph{i.e.,}\xspace}
\newcommand{\eg}{\emph{e.g.,}\xspace}

\newcommand{\RAGE}{\textsc{RAGE}\xspace}

\begin{document}
\fancyhead{}

\settopmatter{printacmref=false, printfolios=false}

\title{Review-Driven Answer Generation for Product-Related Questions in E-Commerce}\thanks{$^\star$Chenliang Li is the corresponding author.}
\author{Shiqian Chen$^1$, Chenliang Li$^{1\star}$, Feng Ji$^2$, Wei Zhou$^2$, Haiqing Chen$^2$}
\affiliation{%
  \institution{
  1. Key Laboratory of Aerospace Information Security and Trusted Computing, Ministry of Education, School of Cyber Science and Engineering, Wuhan University, China
  \\\{sqchen, cllee\}@whu.edu.cn\\ 
  2. Alibaba Group, Hangzhou, China\\\{zhongxiu.jf,fayi.zw,haiqing.chenhq\}@alibaba-inc.com}
}

\begin{abstract}
The users often have many product-related questions before they make a purchase decision in E-commerce. However, it is often time-consuming to examine each user review to identify the desired information. In this paper, we propose a novel \textbf{r}eview-driven framework for \textbf{a}nswer \textbf{g}eneration for product-related questions in \textbf{E}-commerce, named \RAGE. We develope \RAGE on the basis of the multi-layer convolutional architecture to facilitate speed-up of answer generation with the parallel computation. For each question, \RAGE first extracts the relevant review snippets from the reviews of the corresponding product. Then, we devise a mechanism to identify the relevant information from the noise-prone review snippets and incorporate this information to guide the answer generation. The experiments on two real-world E-Commerce datasets show that the proposed \RAGE significantly outperforms the existing alternatives in producing more accurate and informative answers in natural language. Moreover, \RAGE takes much less time for both model training and answer generation than the existing RNN based generation models.
\end{abstract}



\copyrightyear{2019} 
\acmYear{2019} 
\setcopyright{acmcopyright}
\acmConference[WSDM '19]{The Twelfth ACM International Conference on Web Search and Data Mining}{February 11--15, 2019}{Melbourne, VIC, Australia}
\acmBooktitle{The Twelfth ACM International Conference on Web Search and Data Mining (WSDM '19), February 11--15, 2019, Melbourne, VIC, Australia}
\acmPrice{15.00}
\acmDOI{10.1145/3289600.3290971}
\acmISBN{978-1-4503-5940-5/19/02}

\maketitle

{\fontsize{8pt}{8pt} \selectfont
\textbf{ACM Reference Format:}\\
Shiqian Chen, Chenliang Li, Feng Ji, Wei Zhou, Haiqing Chen. 2019.
Review-Driven Answer Generation for Product-Related Questions in ECommerce.
In The Twelfth ACM International Conference on Web Search and
Data Mining (WSDM ’19), February 11–15, 2019, Melbourne, VIC, Australia.
ACM, New York, NY, USA, 9 pages. https://doi.org/10.1145/3289600.3290971
}

\section{Introduction}\label{sec:intro}
\begin{figure}[t]
\centering
\includegraphics[height=55mm,width=0.98\columnwidth]{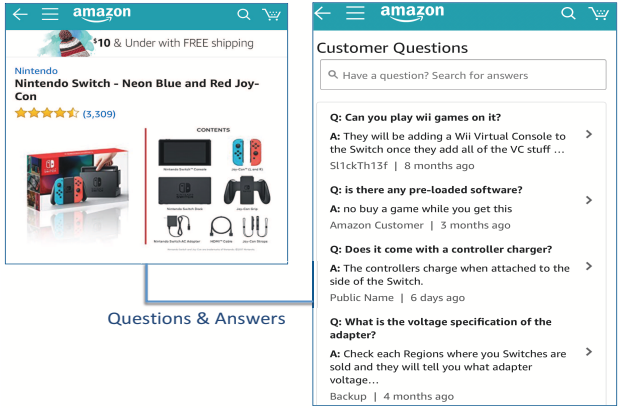}
\caption{Product-related questions and answers for ``Nintendo Switch'' in the auxiliary CQA system of Amazon.}
\label{fig:amazon}
\end{figure}

When people make a purchase offline, questions about diverse aspects of the product are often issued before they make the final decision. The answers for these product-related questions are also sought when they start to purchase in E-Commerce websites. Instead of a face-to-face communication with a salesman, people resort to seeking the useful information from the customer reviews made by the others. These reviews are a fruitful resource where both the characteristics of the product and the opinions from many users can be well revealed. However, browsing each review written by the previous users to identify the product-related information is a time consuming process. To address this demanding information need, leading E-Commerce websites such as Amazon\footnote{\url{http://www.amazon.com}} and Taobao\footnote{\url{http://www.taobao.com}} have established an online community question answering (CQA) system. It allows a user to answer the questions posted by others. Figure~\ref{fig:amazon} illustrates a APP snapshot that contains several questions and their answers for the product ``Nintendo Switch'' in the CQA system of Amazon. These questions cover many aspects of the product, \eg hand controller, power adapter. A restriction is applied in these auxiliary CQA systems, mainly for the purpose of quality guarantee. Only the users that have purchased this product are allowed to answer the product-related questions. Although these CQA systems largely facilitate the information seeking for many users, a new question often takes a long time to receive a response from a consumer. Consequently, a user may still choose to wade through the reviews, due to this \textit{low recall} problem. That is, the answered questions could only cover a small proportion of the product-related questions that may be asked about for a product.

Many aspect-based opinion mining techniques have been developed to mitigate this information overload~\cite{emnlp16:laddha,www16:wang,aaai17:wang}. However, the relevant information is mainly expressed in a bag-of-words (BOW) fashion by the existing solutions. It is often hard for users to precisely interpret these results. It seems that extracting some review sentences containing the relevant information about a product-related question is a promising avenue~\cite{acl16:yan,acl17:cui,acl17:qiu}. However, since no restriction is given on its writing styles, user reviews often contain a lot of syntax errors and lack proper usage of punctuations, making sentence extraction a much difficult task. It is not user-friendly to display an unnatural sentence that contains many irrelevant information. Recently, the notable accomplishments in language generation have reshaped the way that people communicate with computers. The recurrent neural networks (RNNs) based sequence-to-sequence (Seq2Seq) architecture enables us to adopt natural text generation in a scalable and end-to-end manner. Many applications such as dialogue systems~\cite{acl15:shang,ijcai16:yin,emnlp16:li,acl17:qiu,aaai17:xing,acl18;liu,www18:chen}, document summarization~\cite{aaai17:nallapati} and poem generation~\cite{emnlp14:zhang,acl17:zhang} have been developed in the paradigm of Seq2Seq architecture.

\begin{figure}[t]
\centering
\includegraphics[height=40mm,width=0.98\linewidth]{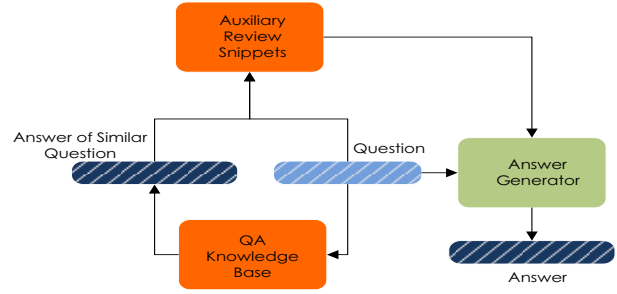}
\caption{The workflow of the proposed \RAGE.}
\label{fig:workflow}
\end{figure}

Motivated by these promising progress, in this paper, we attempt to generate a natural answer for a product-related question based on the relevant information provided in the reviews of that product. Specifically, we propose a novel \textbf{r}eview-driven framework for \textbf{a}nswer \textbf{g}eneration for product-related questions in \textbf{E}-commerce, named \RAGE. Response time is an important factor in E-Commerce. To shorten the response time for answer generation, \RAGE is built on the basis of the recently proposed convolutional Seq2Seq architecture~\cite{icml17:gehring}. The inherent convolutional network structure enables us to apply parallel computation of convolutional operations. The overall workflow of \RAGE is illustrated in Figure~\ref{fig:workflow}. For each question, \RAGE first extracts the review snippets that contain relevant information regarding the question (called \textit{auxiliary review snippets}). To tackle the adverse impact of the noise-prone nature of user reviews, we propose a weighting strategy to highlight the relevant words appearing in the auxiliary review snippets. Then, we propose a method to inject the relevant information provided by the auxiliary review snippets to guide the answer generation. Specifically, we utilize both the attention and gate mechanism to highlight the relevant part presented in the auxiliary review snippets against the question context. Our experiments demonstrate the effectiveness and efficiency of \RAGE in generating accurate and informative answers in natural language for product-related questions in E-Commerce. In summary, the contributions of this paper are three-fold: 

\begin{enumerate}
\item We propose and formalize a new task of review-driven answer generation for product-related questions in E-Commerce domain. The purpose is automatically generate an accurate and informative answer in natural language for a product-related question based on the corresponding user reviews.

\item We propose a noise-tolerant solution based on convolutional neural networks to generate natural answers for product-related questions. To the best of our knowledge, this is the first attempt to generate natural answers for product-related questions based on the unstructured and noisy user reviews in E-Commerce;

\item We conduct extensive experiments on two real-world datasets from a leading E-Commerce service. The experimental results show that \RAGE produces more accurate and informative answers in natural language for product-related questions than the existing alternative techniques, in terms of both automatic evaluation and human examination.
\end{enumerate}

\section{The Proposed Framework}\label{sec:alg}
\begin{figure}[t]
\centering
\includegraphics[height=56mm,width=0.88\linewidth]{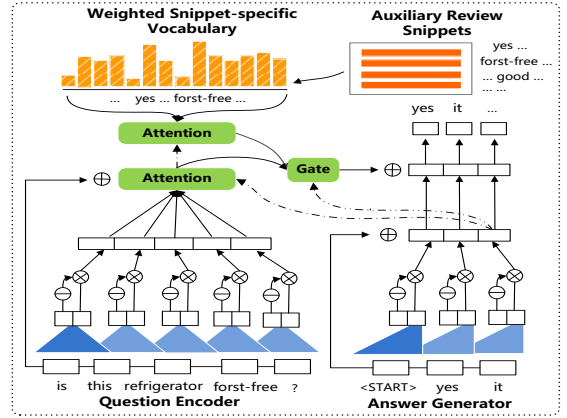}
\caption{The structure of encoder and generator in \RAGE.}
\label{fig:answergenerator}
\end{figure}

In this section, we present the proposed \RAGE for reivew-driven answer genearation in detail. The overall structure of \RAGE is illustrated in Fig~\ref{fig:answergenerator}.

\subsection{Question Encoder}\label{ssec:encoder}
 Given a question $q=(x_1,...,x_m)$, we first construct the dense vector representation for each word $x_i$ by using three embedding vectors:
\begin{align} 
\mathbf{e}_{x_i}=\mathbf{w}_{x_i}+\mathbf{t}_{x_i}+\mathbf{p}_i
\label{eqn:wordrepresentation}
\end{align}
where $\mathbf{w}_{x_i}$ is the pre-trained word embedding for $x_i$, $\mathbf{t}_{x_i}$ is the embedding for the POS tag of word $x_i$, and $\mathbf{p}_i$ is the embedding for the absolute position $i$. Here, we exploit the POS tags of the words to enhance the natural answer generation, since the POS tags could provide the syntactic structure for a sentence and grammatical role of each word. The absolute position embeddings enable the model to be aware of the portion of the question it is currently dealing with. Then, with a sliding window of size $k$, we utilize a multi-layer gated convolutional network~\cite{icml17:dauphin} to extract the hidden state for each word as follow:
\begin{align}
\mathbf{z}^1_i&=\nu(\mathbf{W}^1[\mathbf{e}_{x_{i-k/2}},...,\mathbf{e}_{x_{i+k/2}}]+\mathbf{b}^1)+\mathbf{e}_{x_i}\nonumber\\
&\cdots\label{eqn:encoder}\\
\mathbf{z}^{l_q}_i&=\nu(\mathbf{W}^{l_q}[\mathbf{z}^{l_q-1}_{i-k/2},...,\mathbf{z}^{l_q-1}_{i+k/2}]+\mathbf{b}^{l_q})+\mathbf{z}^{l_q-1}_i\nonumber
\end{align}
where $\mathbf{z}^l_i$ is the hidden state calculated by $l$-th layer for word $x_i$, $\mathbf{W}^{l}$ and $\mathbf{b}^l$ are the parameter matrix and bias vector for the convolutional operation at $l$-th layer, $\nu$ is the nonlinear activitation function. In this work, we restrict the dimension sizes of word embeddings and hidden layers to be identical for simplification. In Equation~\ref{eqn:encoder}, function $\nu()$ is chosen to be \textit{gated linear units} (GLU)~\cite{icml17:dauphin}: 
\begin{align}
\nu([\mathbf{a},\mathbf{b}])=\mathbf{a}\odot\sigma(\mathbf{b})
\end{align}
where $\odot$ denotes the element-wise product of two vectors, $\sigma$ is the sigmoid function. GLU works as a gating mechanism to identify the relevant features extracted by the convolution operation. Here, we stack $l_q$ layers on top of each other and hence create a hierarchical structure over the whole question. Because the higher layers consider more distant words through the hidden states encoded by the lower layers, a much larger context around each word is represented in the last layer. That is, the long-range dependencies can be modeled in this hierarchical structure. The residual connection is added between two consecutive layers to facilitate the deep network learning. Both $k/2$ zero vectors are padded to the begining and the end of the input at each layer. We apply a linear transformation to the hidden states at the last layer as the output of the question encoder: 
\begin{align}
\mathbf{z}_i=\mathbf{W}_e\mathbf{z}^{l_q}_i+\mathbf{b}_e
\end{align}
where $\mathbf{z}_i$ is the encoder output for word $x_i$ in the question.

\subsection{Review-Driven Answer Generator}\label{ssec:decoder}
\paratitle{Basic Answer Generator.} Similar to the question encoder, we also utilize a multi-layer gated convolutional network for answer generation. Here, we introduce a special symbol $y_0$ indicating the start of the answer. After generating $j$-th words by step $j$, we have an answer sequence $T=(y_0,y_1,...,y_j)$. Given the answer generator has $l_a$ gated convolutional layers,  the hidden state for word $y_j$ at each layer is then hierarchically calculated as follows:
\begin{align}
\mathbf{h}^1_j&=\nu(\mathbf{W}^1_a[\mathbf{e}_{y_{j-k/2}},...,\mathbf{e}_{y_j}]+\mathbf{b}^1_a)+\mathbf{e}_{y_j}\nonumber\\
&\cdots\label{eqn:decoder}\\
\mathbf{h}^{l_a}_j&=\nu(\mathbf{W}^{l_a}_a[\mathbf{h}^{l_a-1}_{j-k/2},...,\mathbf{h}^{l_a-1}_j]+\mathbf{b}^{l_a}_a)+\mathbf{h}^{l_a-1}_j\nonumber
\end{align}
where $\mathbf{h}^l_j$ is the hidden state calculated by $l$-th layer for word $y_j$, $\mathbf{W}^l_a$ and $\mathbf{b}^l_a$ are the parameter matrix and bias vector for the convolutional operation at $l$-th layer. The generation probability for the next word $y_{j+1}$ is then calculated through a linear transformation and a softmax function: 
\begin{align}
p(y_{j+1}|y_0,...,y_j,q)=\text{softmax}(\mathbf{W}_o\mathbf{h}^{l_a}_{j}+\mathbf{b}_o).\label{eqn:nextprob}
\end{align}
where $\mathbf{h}^{l_a}_{j}$ denotes the hidden state calculated at the last layer for word $y_j$. Note that a word representation for the answer generator is also formed by three constituents: word embedding, POS embedding and position embedding. Here, we take the dominating POS tag of each word within the collection as input during the answer generation. 

Actually, the hidden state calculated in each layer incorporates the question context information through a multi-step attention mechanism. A separate attention mechanism is introduced for each layer in answer generator. Specifically, given a hidden state $\mathbf{h}^l_j$, we first apply a linear transformation and combine the result with the embedding vector of word $y_j$: $\mathbf{d}^l_j=\mathbf{W}^l_d\mathbf{h}^l_j+\mathbf{b}^l_d+\mathbf{e}_{y_j}$. Then the attentive weight for each word in the question is calculated with a softmax function:
\begin{align}
a^l_{i,j}=\frac{\exp(\mathbf{d}_j^{l}\cdot\mathbf{z}_i)}{\sum_k\exp(\mathbf{d}_j^{l}\cdot\mathbf{z}_k)}\label{eqn:qatt}
\end{align}
where $\cdot$ denotes the dot-product, $a^l_{i,j}$ is the weight indicating the relatedness between the contextual information around question word $x_i$ and answer word $y_j$ in $l$-th layer. The question context $\mathbf{c}^l_{j}$ for word $y_j$ is then computed as follows:
\begin{align}
\mathbf{c}^l_{j}=\sum_i a^l_{i,j}(\mathbf{z}_i+\mathbf{e}_{x_i})\label{eqn:qc}
\end{align}
In Equation~\ref{eqn:qc}, both the encoder outputs and the corresponding word representations are considered in the context. The inclusion of $\mathbf{e}_{x_i}$ is found to be beneficial in~\cite{icml17:gehring}. At last, we update $\mathbf{h}^l_j$ through the addition of $\mathbf{c}^l_{j}$: $\mathbf{h}^l_j=\mathbf{h}^l_j+\mathbf{c}^l_{j}$. The updated $\mathbf{h}^l_j$ is fed into the next layer as input for further hidden state calculation.

\paratitle{Incorporating Review Guidance.} For each question, we first need to extract the review snippets that contain the relevant information from the user reviews of the corresponding product (called \textit{auxiliary review snippets}). The detailed auxiliary review snippet extraction process is described in Section~\ref{ssec:data}. Suppose $n_q$ auxiliary review snippets $\mathbb{S}_q=\{s_1,...,s_{n_q}\}$ are extracted for question $q$. Although each auxiliary review snippet would contain some noisy information, the relevant information becomes obvious when we consider these auxiliary review snippets together. That is, we believe that the dominating information expressed in these auxiliary review snippets will be the core content to answer the question.  We choose to weight the words appearing in these snippets based on their frequency and semantic relatedness. Specifically, let $\mathbb{V}_q$ be the snippet-specific vocabulary for the snippet set $\mathbb{S}_q$, we calculate weight $\omega_{r_i}$ for each word $r_i$ inside $\mathbb{V}_q$ as follows:
\begin{align}
\omega_{r_i}=\frac{f_s(r_i,\mathbb{S}_q)}{n_q}\sum_{s\in\mathbb{S}_q}\text{rel}_{max}(r_i,s)
\end{align}
where $f_s(r_i,\mathbb{S}_q)$ returns the number of the snippets in $\mathbb{S}_q$ that contain word $r_i$, $\text{rel}_{\text{max}}(r_i,s)$ returns the maximum cosine similarity between word $r_i$ and the words in snippet $s$ in terms of word embeddings. A word with higher snippet frequency and highly relevant to each snippet will receive a larger weight. We normalize $\omega_{r_i}$ by using max-normalization: $\hat{\omega}_{r_i}=\omega_{r_i}/\max_{r_j\in\mathbb{V}_q}(\omega_{r_j})$. The importance for each word in the vocabulary $\mathbb{V}_q$ is then encoded by magnifying the corresponding word embeddings with the weights: $\hat{\mathbf{w}}_{r_i}=\hat{\omega}_{r_i}\mathbf{w}_{r_i}$.

We then utilize the attention and gate mechanisms to inject the relevant information highlighted in the reviews into the answer generation process.
\begin{align}
a^{l,r}_{j,i}&=\frac{\exp(\mathbf{c}_j^{l}\hat{\mathbf{w}}_{r_i})}{\sum_k\exp(\mathbf{c}_j^{l}\hat{\mathbf{w}}_{r_k})}\label{eqn:ratt}\\
\mathbf{o}^{l}_{j}&=\sum_i a^{l,r}_{j,i}\hat{\mathbf{w}}_{r_i}\\
g^{l,r}_j&=f^l(\mathbf{h}_j^l,\mathbf{c}^l_j,\mathbf{o}^{l}_{j})\\
\mathbf{h}_j^l&=\mathbf{h}_j^l+g^{l,r}_j\cdot \mathbf{c}_j^l+(1-g^{l,r}_j)\cdot \mathbf{o}^{l}_{j}
\end{align}
where $f^l$ is a neural network function with a two-layer MLP and works as a gate to control the impact of $\mathbf{o}^{l}_j$ for the final decoder state $h_i^l$ at $l$-th layer. Note that the attentive weight $a^{l,r}_{j,i}$ is calculated based on the question context $\mathbf{c}_j^l$ in Equation~\ref{eqn:ratt}. Recall that the question context $\mathbf{c}_j^l$ is also calculated through an attention mechanism based on the generated word $y_j$. In this sense, $\mathbf{o}^{l}_j$ encodes the relevant information provided in the auxiliary review snippets by considering both the question and the partially generated answer at the current step.

\section{Experiment}\label{sec:exp}
In this section, we evaluate the proposed \RAGE\footnote{\scriptsize The implementation is available at~\url{https://github.com/WHUIR/RAGE}.} and existing state-of-the-art alternative generation models over two real-world E-Commerce QA datasets with user reviews.
\eat{Firstly, we introduce the two datasets used in the evaluation and the procedure for auxiliary review snippet extraction in Section~\ref{ssec:data}. Then, we describe the experimental settings in Section~\ref{ssec:settings}, including the baseline models and evaluation metrics. At last, we report the experimental results and make discussion in Section~\ref{ssec:comparison}
}

\subsection{Data Preparation}\label{ssec:data}
We construct two real-world datasets with $4,457$/$47,979$ products under category \textit{Cellphone}/\textit{Household Electrics} respectively in Taobao, a leading E-Commerce website in China. Note that category \textit{Household Electrics} is much larger and more diverse since it includes many subcategories like \textit{Television}, \textit{Refrigerator}, \textit{Kitchen Appliances}. The associated question answer (QA) pairs and user reviews for these products are included. Note that a question for a specific product may have more than one answer. Here, we just pick the longest answer to form a QA pair. We reserve the products of $6$ brands and $2$ subcategories from \textit{Cellphone} and \textit{Household Electrics} respectively for test set building, where the rest are taken as the training sets. The two test sets are built by sampling about $1,000$ QA pairs from these reserved products. We perform the following preprocessing steps on the two datasets: (1) perform Chinese word segmentation and POS tagging for these QA pairs and reviews with a proprietary Chinese NLP toolkit; (2) remove all reviews with fewer than 10 words; (3) remove QA pairs that have fewer than $4$ words in either question or answer part; (4) remove QA pairs that have more than $40$ words in either question or answer part. The word embeddings are then learnt over the combination of the rest QA pairs and reviews using Goolge's Word2Vec toolkit\footnote{\scriptsize \url{https://code.google.com/p/word2vec}} with Skip-gram algorithm~\cite{arxiv13:mikolov}.

\paratitle{Auxiliary Review Snippet Extraction.} We need to extract the auxiliary review snippets that contain the information relevant to the question from the user reviews. Word Mover's Distance (WMD)~\cite{icml15:kusner} is an effective metric to measure the semantic similarity between two short documents based on the word embeddings. Instead of relying on keyword matching like cosine similarity, WMD attempts to find an optimal traveling cost between two documents in the word embedding space. WMD has been widely adopted in the literature to measure short text semantic similarity~\cite{icse16:ye,cikm17:julia}. Here, we adopt WMD as the relevance measure for a pair of two documents. For a training QA pair, we can directly concatenate the question with its answer together to perform auxiliary review snippet extraction. However, for a testing QA pair, only the question is available. In this case, we derive a question expansion similar to the query expansion in information retrieval. For question $q$ of a testing QA pair,  we search the answered questions of other products in the training set. We retrieve the most relevant question $q'$ using WMD. Question $q$ is then augmented with the answer of question $q'$. It is intuitive that the aspect or opinion words contained in these answers help identify the relevant information from the reviews.

For each review, a sliding window of size $t$ is utilized to extract a review snippet that is most relevant to a QA pair. Here, we set $t$ to be $10$ words in this work. The most relevant review snippet from each review is retained for further processing. Then, we filter out these review snippets with a threshold $\pi$ in terms of WMD measure. The left review snippets are considered as the auxiliary review snippets for this QA pair. After manual examination over a sample of $100$ QA pairs, we find that top relevant review snippets for a QA pair largely contain the information relevant to the answer. Hence, we set threshold $\pi$ to be the averaged WMD score of all highly relevant review snippets of all QA pairs. A review snippet is considered to be highly relevant if it is among top-$10$ relevant review snippets for the corresponding QA pair. Based on this $\pi$ value, we can obtain a set of auxiliary review snippets for each QA pair. We then filter out the QA pairs with less than two auxiliary review snippets. Table~\ref{tbl:stats} summarizes the detailed statistics about the resultant training and test sets.

\begin{table}
\small
\centering
\caption{\label{tbl:stats} The statistics about \textit{Cellphone} and \textit{Household Electrics} dataset.}
\begin{tabular}{p{0.46in}|c||cccc}
\toprule
 Dataset & Set & $\#(q,a)$ & Avg($|q|$) & Avg($|a|$) & Avg($|\mathbb{S}_q|$) \\
\hline
  \multirow{2}{0.46in}{Cellphone} & Train & 355,986 & 9 & 13 & 7 \\
  & Test & 856 & 8 & 13 & 6 \\
\midrule
 \multirow{2}{0.46in}{Household Electrics} & Train & 797,688 &11 & 12 & 9 \\
  & Test & 1,000 & 11 & 12 & 9 \\
\bottomrule
\end{tabular}
\end{table}

\subsection{Experimental Settings}\label{ssec:settings}

\paratitle{Baselines.} Since the proposed \RAGE is the first model for answer generation based on the relevant review information in E-Commerce, there is no previous works that can be directly applied for the task. We choose several alternative techniques for performance comparison:

\begin{description}

\item[S2SA] The standard Seq2Seq model with attention mechanism~\cite{corr14:bahdanau}. The RNN encoder and decoder are instantiated using gated recurrent unit (GRU)~\cite{corr14:chung}. We utilize a bidirectional GRU to encode the question. Since no product-related review information is provided, S2SA only generates the answer based on the information provided in the training set.

\item[TA-S2S] The topical words relevant to the question are taken as the topical words for the topic-aware sequence-to-sequence (TAS2S) model~\cite{aaai17:xing}. This can be considered as an adaptation of TAS2S for review-driven answer generation. Following the work in~\cite{aaai17:xing}, we firstly assign a topic to the auxiliary review snippets of the given question, based on a Twitter-LDA~\cite{ecir11:zhao} model learnt over all the reviews in the training set. Then the top-$100$ topical words under the same topic are incorporated into the decoder for answer generation.

\item[ConvS2S] The convolutional sequence to sequence model (ConvS2S) proposed in~\cite{icml17:gehring}. The proposed \RAGE is built on the basis of ConvS2S. As with S2SA, no corresponding review information is provided for answer generation;

\item[ConvS2S-RV] The auxiliary review snippets are provided for ConvS2S during the answer generation. When generating an answer, we directly restrict the inference vocabulary to the words that appear in the auxiliary review snippets. This can be considered as an adaptation of a dynamic vocabulary sequence-to-sequence model~\cite{aaai18:wu} for review-driven answer generation.

\item[RAGE/POS] The proposed \RAGE model without exploiting the POS tags (ref. Equation~\ref{eqn:wordrepresentation}).

\end{description}

\paratitle{Evaluation Metrics.} We adopt both automatic evaluation and human evaluation to measure the performance of answer generation. For each method in comparison, we only take the top response generated by beam search as the answer for performance evaluation.

\textbf{Distinct-1/2:} Distinct-1 and distinct-2 calculate the ratios of distinct unigrams and bigrams in generated response~\cite{naacl16:li,aaai17:xing,aaai18:wu}. This is a widely used metric for response generation. The higher score suggests that the generated response is more diverse. Here, we report the averaged answer-level distinct-1 and distinct-2 scores over all the testing questions.

\textbf{Embedding based Similarity (ES):}~\cite{emnlp16:liu} show that averaged embedding based cosine similarity correlates well with human judgment to measure relevance between two responses. A larger ES score suggests that the generated answer is more similar to the ground truth, which indicates better generation performance.

\begin{CJK}{UTF8}{gbsn}
\textbf{Human Annotation:} We further recruit two human annotators to judge the quality of the generated answer for a random sample of $400$ test QA pairs for each dataset. These two annotators have rich knowledge in both Taobao E-Commerce platform and the products of cellphone/household electrics category. Answers generated by all the methods are pooled and randomly shuffled for each annotator. The ground truth answer and the associated auxiliary review snippets are also provided for each question to facilitate the human annotation. We also provide the URL for the corresponding product of the test QA pair to encourage the annotators to search more relevant information from the user reviews for better judgment. Note that each generated answer is evaluated by the both annotators.  A score between 0 and 3 is assigned to each generated answer based on the following criteria: \textbf{+3}: the answer is not only natural and relevant to the question, but also informative and accurate according to the ground truth answer and user reviews; \textbf{+2}: the answer is informative and accurate, but is not completely natural and error-prone in grammar; \textbf{+1}: the answer can be used as a reply, but is not informative enough (\eg ''不错'' (not bad), ''可以买'' (you can buy), ''我觉得还可以'' (it is ok for me)); \textbf{0}: the answer is irrelevant and unclear in meaning (\eg too many grammatical errors to understand). To ease the interpretation for annotation scores, we also report the averaged score for each method over the chosen instances. A higher score suggests a better performance.

Here, we do not choose BLEU~\cite{acl02:papineni} for automatic evaluation. The BLEU measures the generated answer against the ground truth answer in terms of word matching. However, a Chinese word often consists of more than one single character. Hence, the semantic similarity is often not well captured by counting the word overlap. Moreover, the same meaning can be expressed in totally different ways. Using BLEU may not count the language diversity. For example, given a ground truth answer ''老妈用起来挺好'' (Old mummy has a nice user experience) for quesiton ''这款手机适合老人用么?'' (Is this product suitable for the old?), the generated answer would be ''非常适合老人'' (very appropriate for the old). Though there is no word overlap between the ground truth answer and the generated one, apparently they convey the same semantic meaning. Therefore, we should consider the generated answer to be correct. However, using BLEU would give this case a negative judgement.

\end{CJK}

\paratitle{Parameter Setting.} Each generation model is trained with a batch size of $64$ and a dimension size of $300$ for word, POS and position embeddings. A random sample of $1,000$ QA pairs in the training set is used as the validation set for early stop and parameter selection. Specifically, for RNN based Seq2Seq models, the hidden dimension size is set to be $1000$. For convolutional models, we use a four-layer gated convolutional network for both encoder and generator (\ie $l_q=l_a=4$). The hidden dimension size of hidden layers is $300$. Following the setting recommended in~\cite{icml17:gehring}, the convolutional models are trained with Nesterov's accelerated gradient method~\cite{icml13:sutskever}, and the window size $k$ is set to be $2$ and $4$ for the encoder and generator respectively. We also add L2 regularization term with the coefficient of $0.001$ for all generation models. There are $54$/$55$ distinct POS tags for \textit{Cellphone} and \textit{Household Electrics} respectively. The word embeddings are pre-trained using Google's Word2Vec toolkit as mentioned previously, and fixed for the model training. The POS embeddings and position embeddings are fine-tuned during the training.

\begin{table}
\small
\centering
\caption{\label{tbl:autometrics} Automatic evaluation on two datasets. The best and second best results are highlighted in boldface and underlined respectively.}
\begin{tabular}{p{0.46in}|l||ccc}
\toprule
 Dataset & Model & distinct-1 & distinct-2 & ES  \\
\hline\hline
  \multirow{6}{0.46in}{Cellphone} & S2SA & 0.1568 & 0.4685 & 0.5323 \\
  & TA-S2S & 0.2349 & 0.4384 & 0.6062\\
  & ConvS2S & 0.1662 & 0.3769 & 0.5677 \\
  & ConvS2S-RV & 0.1857 & 0.3558 & 0.6008 \\
  & RAGE/POS &\underline{0.2391} & \underline{0.4915} & \underline{0.6418}\\
  & RAGE & \textbf{0.2407} & \textbf{0.5135} & \textbf{0.6505} \\
  \midrule
  \multirow{6}{0.46in}{Household Electrics} & S2SA & 0.2635 & 0.3671 & 0.5399 \\
  & TA-S2S & 0.2797 & 0.4864 & 0.5500 \\
  & ConvS2S & 0.2646 & 0.4698 & 0.4801 \\
  & ConvS2S-RV & 0.2240 & 0.5135 & 0.5606 \\
  & RAGE/POS & \textbf{0.5156} & \textbf{0.6187} & \underline{0.5790}\\
  & RAGE & \underline{0.3917} & \underline{0.5290} & \textbf{0.5880} \\
\bottomrule
\end{tabular}
\end{table}

\subsection{Performance Comparison}\label{ssec:comparison}
\begin{table*}
\small
\centering
\caption{\label{tbl:human} Performance comparison of different models on two datasets in terms of human evaluation. The best and second best results are highlighted in boldface and underlined respectively (excluding GroundTruth). GroundTruth: the answer generated by the consumer in the auxiliary QA system.}
\begin{tabular}{l||cccccc|cccccc}
\toprule
& \multicolumn{6}{c|}{Cellphone} & \multicolumn{6}{c}{Household Electrics}\\
 \cline{2-13}\raisebox{1.5ex}[0pt]{Model} & 0 & 1 & 2 & 3 & Avg & Kappa & 0 & 1 & 2 & 3 & Avg & Kappa\\
\hline\hline





  S2SA & \underline{56.25}$\%$ & \underline{31.50}$\%$ & $5.75\%$ & 6.50$\%$ & 0.6250& 0.5081 &45.25$\%$ &\textbf{46.00}$\%$ &$6.00\%$ &$2.75\%$ &0.6625 & 0.4992 \\

  TA-S2S & $51.00\%$ & $17.25\%$ & \underline{21.50}$\%$ & \underline{10.25}$\%$ &0.9100 &0.5134 &\textbf{51.00}$\%$ &\underline{36.50}$\%$ &$6.25\%$&$6.25\%$ &0.6775 &0.5706 \\

  ConvS2S & $54.75\%$ & $30.75\%$ & 10.50$\%$ &4.00$\%$ & 0.6375 & 0.5533 &$43.50\%$ &$24.75\%$ &\textbf{14.25}$\%$ & 17.50$\%$ & 1.0575 & 0.5707 \\

  ConvS2S-RV & \textbf{56.50}$\%$ & 31.00$\%$ & $9.00\%$ & $3.50\%$ & 0.5950 & 0.5673 & \underline{45.75}$\%$ & 30.25$\%$ & $10.00\%$ & $14.00\%$ & 0.9225 & 0.5971 \\

  RAGE/POS & $35.50\%$ & \textbf{38.00}$\%$ & 16.75$\%$ & 9.75$\%$ & \underline{1.0075} & 0.6019 & 41.00$\%$& 30.50$\%$ & 9.75$\%$ & \underline{18.75}$\%$ &\underline{1.0625} &0.5528 \\

  RAGE & $35.25\%$ & $22.00\%$ & \textbf{23.25}$\%$ & \textbf{19.50}$\%$ &\textbf{1.2700} & 0.6174 & $25.75\%$ &$30.00\%$ & \underline{13.00}$\%$ &\textbf{31.25}$\%$ &\textbf{1.4975} & 0.5974 \\

  GroundTruth & $14.00\%$ & $22.25\%$ & 9.25$\%$ & 54.5$\%$ & 2.0425 & 0.6589 & $5.50\%$ & $20.50\%$ & 13.25$\%$ & 60.75$\%$ & 2.2925 & 0.6278 \\
\bottomrule
\end{tabular}
\end{table*}

\paratitle{Automatic Metrics.} Table~\ref{tbl:autometrics} reports the performance comparison in terms of automatic metrics. Here, we make the following observations:

First, while all the models achieve relatively higher distinct-1/2 performance on \textit{Household Electrics}, a relatively lower ES performance, however, is achieved by all the models. This is reasonable because \textit{Household Electrics} is more diverse than \textit{Cellphone}. It is more likely to generate a diverse response by training the generator over a diverse training set. On the contrary, it is more difficult to learn the question-answer patterns over the same dataset, leading to a lower ES performance.

Second, the proposed \RAGE and \RAGE/POS significantly outperform the baseline methods in terms of distinct-1/2 and ES metrics. Another observation is that \RAGE/POS achieves much better distinct-1/2 performance than \RAGE on \textit{Household Electrics}. However, further human evaluation reveals that more answers generated by \RAGE/POS contain defects. Many irrelevant words are generated by \RAGE/POS, leading to higher distinct-1/2 scores instead. This suggest that incorporting the POS information is beneficial for generating accurate and informative answers.

Third, ConvS2S-RV significantly outperforms ConvS2S method in terms of ES metric. Indeed, by restricting the inference vocabulary to the words inside the auxiliary review snippets could enhance the similarity against the ground truth answer by generating more relevant words regarding the question. Similarly, TA-S2S also performs better than vanilla S2SA under ES metric. These observations partially confirm that incorporating the guidance of the auxiliary review snippets can produce more relevant answers, compared to the ground truth answers generated by the consumers.

\paratitle{Human Annotation.} Table~\ref{tbl:human} reports the performance comparison by human annotation. Given that the ground truth answers are provided by human consumers, we also choose to report the evaluation by human annotation on these human generated answers as a reference. Cohen's kappa coefficient of the human annotations is also included in Table~\ref{tbl:human} for each method. Here, we make the following observations.

First, it is obvious that \RAGE generates more accurate, informative and acceptable natural answers (\ie answers with score 3 or 2), and much less invalid answers (\ie answers with score 0) than the baseline methods. This observation is consistent with the results from the automatic evaluation. It is interesting to see that ConvS2S-RV achieves inferior performance than ConvS2S in terms of human evaluation. In contrast, the proposed \RAGE gains a much large improvement against ConvS2S. This validates the effectivenss of the proposed mechanism to incorporate the auxiliary review snippets for better answer generation. Note that \RAGE outperforms \RAGE/POS with a large margin here. This also validates the effectiveness of incorporating POS information for better natural answer generation.

Second, TA-S2S performs better than vanilla S2SA in terms of averaged annotation score. This observation is consistent with the result of automatic evaluation. Furthermore, there is still a substantial performance gap between the human performers and automatic answer generators. Even \RAGE achieves much lower performance than the consumers (\ie GroundTruth). This suggests that review-driven answer generation is a very challenging task. Around $55\%$ and $61\%$ of the human generated answers are accurate and informative for \textit{Cellphone} and \textit{Household Electrics} respectively. However, there are also about $36\%$ and $26\%$ of the human generated answers are also meaningless or uninformative (\ie answers with score 0 or 1) respectively.

Third, we find that the kappa coefficient generally correlate well with the human evaluation performance (\ie in terms of averaged annotation score). This is reasonable because it is easy to reach agreement on the accurate and informative answers. Further human examination reveals that the majority of disagreements in human annotation happens when the scores 0 and 1 are assigned. That is, it is more difficult to separate the general answers from invalid ones. The major reason is that most general answers contain many irrelevant information.

\begin{CJK}{UTF8}{gbsn}
\begin{table*}[t]
\small
\centering
\caption{\label{tbl:examples} The answers generated by different models for the sampled questions. MRS: the most relevant auxiliary review snippet. The first three questions are from \textit{Cellphone}, while the rest three are from \textit{Household Electrics.}}
\begin{tabular}{|l||p{5.5in}|}
\toprule
Question~1 & 电池耐用吗？ (Does it have a long battery life?) \\
\midrule
S2SA & 我不知道 (I do not know)\\
\hline
TA-S2S & 还可以,电池 (The battery is ok)\\
\hline
ConvS2S & 我的还可以 (Mine is ok)\\
\hline
ConvS2S-RV & 可以可以,电池可以 (Okok, the battery is ok)\\
\hline
RAGE/POS & 还可以,电池耐用 (Ok, the battery life is long)\\
\hline
RAGE & 还可以,电池耐用 (Ok, the battery life is long)\\
\hline
MRS & 不错的! 电池挺耐用的,声音超大 (great! The battery life is very long, the sound is loud)\\
\hline\hline
Question~2 & 指纹解锁否？(Does it support fingerprint unlock?) \\
\midrule
S2SA & 我的是的 (The cellphone bought by me is ok)\\
\hline
TA-S2S & 可以,可以的 (Yes,it does)\\
\hline
ConvS2S & 我是给别人买的 (It is bought by me for someone)\\
\hline
ConvS2S-RV & 手机,指纹解锁 (Cellphone, fingerprint unlock)\\
\hline
RAGE/POS & 我的是指纹解锁,很好 (Mine supports fingerprint unlock, very good)\\
\hline
RAGE & 可以指纹解锁 (It supports the fingerprint unlock)\\
\hline
MRS & 是指纹解锁,一点就开,一开始  (It is fingerprint unlock, unlock by one press, at the very start) \\
\hline\hline
Question~3 & 通话质量怎么样? 信号效果怎么样? (how good is the conversation quality? How good is the signal quality?)\\
\midrule
S2SA & 还可以,我不知道 (It is ok, I do not know)\\
\hline
TA-S2S & 还可以,声音很大,屏幕大 (It is ok, the sound is loud, and the screen is big )\\
\hline
ConvS2S & 还可以,声音大 (It is ok, the sound is loud)\\
\hline
ConvS2S-RV & 可以,可以,用着还可以 (It is ok, It is ok, the experience is ok) \\
\hline
RAGE/POS & 还可以,声音大,信号也可以 (It is ok, the sound is very loud, the signal is good)\\
\hline
RAGE & 还可以,声音挺大,信号也可以 (It is ok, the sound is very loud, the signal is good)\\
\hline
MRS & 。通话效果还可以,老爸用着特别 (. The conversation quality is good, old father feels especially)\\
\hline\hline
Question~4 & 这款冰箱是无霜的吗 (Is this refrigerator frost-free?)\\
\midrule
S2SA & 还没用,还没使用 (Have not used yet, have not used yet)\\
\hline
TA-S2S & 还可以,我是给别人买的,我也不知道 (It is ok, it is bought by me for someone, I do not know)\\
\hline
ConvS2S & 还没用,还没用 (Have not used yet, have not used yet)\\
\hline
ConvS2S-RV & 我买了,挺好的 (I have gotten one, very nice) \\
\hline
RAGE/POS & 我觉得还可以 (I think it is ok)\\
\hline
RAGE & 我的是无霜的,用了一段时间了,没发现有霜 (Mine is frost-free. I have used it for a while, frosting is not found.)\\
\hline
MRS & 了两个冰柜,无霜的,妈妈非常满意 (Two freezer, frost-free, Mum is pleased)
\\
\hline\hline
Question~5 & 这款用的怎么样,只有制冷效果吗,那么制热呢 (How about this product? does it only perform cooling? how about its heating effect?)\\
\midrule
S2SA & 还可以,制冷效果还不错,制冷效果还不错 (It is ok, good at cooling, good at cooling)\\
\hline
TA-S2S & 还没安装,不知道 (Have not installed yet, I do not know)\\
\hline
ConvS2S & 还不错,制冷效果还不错,制冷效果还不错 (Not bad, good at cooling, good at cooling)\\
\hline
ConvS2S-RV & 还不错,制冷效果还不错,制冷效果还不错 (Not bad, good at cooling, good at cooling) \\
\hline
RAGE/POS & 制热效果不错 (Good at heating)\\
\hline
RAGE & 制冷效果不错,制热效果不错 (Good at cooling, good at heating)\\
\hline
MRS & 制热挺好,制冷还挺不错的 (Goot at heating, good at cooling)
\\
\hline\hline
Question~6 & 这个质量怎么样,安全不,会自动断电么? (How about this product? is it saft? does it support automatic power-off?)\\
\midrule
S2SA & 还没使用,还没问题 (Have not used yet, no problem so far)\\
\hline
TA-S2S & 还可以,就是加热慢 (It is ok, just slow at heating)\\
\hline
ConvS2S & 还没用,还没用 (Have not used yet, have not used yet)\\
\hline
ConvS2S-RV & 质量不错,质量不错 (The quality is good, the quality is good) \\
\hline
RAGE/POS & 还没用,不知道 (Have not used yet, I do not know) \\
\hline
RAGE & 还可以,就是加热有点慢,我用了两个月了,没发现有漏电现象 (It is ok, just a bit slow at heating, I have used it for two months, no electric leakage is found)\\
\hline
MRS & 够了自动断电,热的有点慢 (enough automatic power-off, slow at heating)
\\
\bottomrule
\end{tabular}
\end{table*}
\end{CJK}

\paratitle{Case Study.} We further investigate the generated answers by looking at some specific examples. Three questions from each dataset are picked here for case study. Table~\ref{tbl:examples} lists the answers generated by each model for performance comparison. The most relevant auxiliary review snippet (MRS) for each question is also included as a reference. Question~1, 2 and~4 are simple questions, because each of them refers to a specific product-related aspect. On the contrary, Question~3, 5 and~6 are considered as complex questions since each of them contains two sub-questions.

Here, we make several observations. First, it is clear that the most relevant snippet often contains irrelevant information due to the informal writing style of user reviews. In detail, MRS contains irrelevant information in $5$ out of $6$ questions. Second, it is observed that S2SA and ConvS2S generate meaningless or uninformative answers in most case. This is reasonable since no product-related review information is provided. Hence, it is unrealistic for them to respond accurately. With the guidance from the auxiliary review snippets, TA-S2S, ConvS2S-RV could generate informative and accurate answers for some simple questions. However, they all failed to provide a complete answer to more complex questions. This observation is also made for \RAGE/POS. Specifically, \RAGE/POS only manages to provide an accurate and informative answers for question~1, 2 and~3. Either meaningless or incomplete answers are generated by \RAGE/POS for question~4, 5 and~6. Third, it is also obvious that the answers generated by \RAGE is more accurate, informative and natural. For example, all alternative solutions only provide an incomplete answers that cover $1$ out of $2$ aspects (\ie the cooling quality and the heating quality). However, \RAGE successfully generates an answer that full covers the two aspects. Another example is about the answers generated by \RAGE for question~4 and~6. The sentences like ``I have used it for a while'' or ``I have used it for two months'' are very informative and convincing. \RAGE effectively absorbs the relevant and useful information provided by the auxiliary review snippets for natural answer generation.

\begin{table}
\small
\centering
\caption{The time cost in seconds for one epoch model training (T) and answer generation (G) by different models.}
\label{tbl:efficiency}
\begin{tabular}{l||cc|cc}
\toprule
& \multicolumn{2}{c|}{Cellphone} & \multicolumn{2}{c}{Household Electrics}\\
 \cline{2-5}\raisebox{1.5ex}[0pt]{Model} & T & G & T & G  \\
\hline\hline
  S2SA & 2,940 & 18 & 5,163 & 18\\
  TA-S2S & 3,862 & 19 & 6,936 & 20\\
  ConvS2S & 791 & 7 & 1,923 & 8\\
  ConvS2S-RV & 791 & 13 & 1,923 & 11\\
  RAGE/POS & 908 & 10 & 2,760 & 11\\
  RAGE & 989 & 11 & 4,146 & 13\\
\bottomrule
\end{tabular}
\end{table}

In summary, we have performed the experiments in terms of automatic evaluation, human annotation and case study. The overall comparison demonstrates the effectiveness of \RAGE to incorporate review guidance for natural answer generation for product-related questions in E-Commerce.

\paratitle{Computational Efficiency.} Response time is an important consideration for real-time answer generation in E-Commerce. Here, we report both the training time and the time taken for answer generation by different generation models in Table~\ref{tbl:efficiency}. Specifically, we record the training time per epoch and answer generation time for each model under the same computing environment with a Tesla k40 GPU. For fair comparison, all the models are implemented by using TensorFlow framework. Note that for review-based answer generation, we need to conduct auxiliary review snippet extraction for each testing question. However, there are many ways to speed up this procedure by the parallel computing. We therefore exclude the time for auxiliary review snippet extraction from the consideration. Similarly, for TA-S2S, we omit the time taken for topic inference and topical word extraction. We can see that S2SA and TA-S2S take much more time for model training and answer generation. Apparently, ConvS2S, ConvS2S-RV, \RAGE/POS and \RAGE take much less time for training and generation due to the parallization of the convolutional operations. Overall, the proposed \RAGE obtains promising generation performance in terms of both effectiveness and efficiency.

\section{Related Work}\label{sec:related}
Facilitating information seeking processing in E-Commerce is always an important task that has drawn great attention from the research community. Many works focus on aspect-based extraction and opinion mining from user reviews~\cite{cikm12:moghaddam,ijcai15:liu,aaai17:wang,ijcai17:ma}. The techniques utilized to address this task include neural networks~\cite{aaai17:wang,ijcai17:ma}, topic modeling~\cite{cikm12:moghaddam,emnlp16:laddha,www16:wang} and rule-based mining~\cite{ijcai15:liu}. However, these existing works mainly present the aspect specific information with a bag-of-words representation, such as opinion terms or sentiment polarity. The users often feel hard to interpret these results precisely. Also, many product-related questions may not be addressed directly based on the extracted aspect information.~\cite{www16:julian} proposes a machine learning model to identify the reviews relevant to a product-related question. However, their model still can not provide a natural and concise answer, which we aim to solve in this work. Recently, several chatbot systems have been developed for E-Commerce websites~\cite{acl17:qiu,acl17:cui,wsdm18:yu} by retrieving the most relevant review sentences or answered questions. However, sentence extraction is often problematic for reviews due to the colloquial language of the latter.

Our work is inspired by the recent advances on neural natural langauge dialogue. The Seq2Seq structure is firstly proposed in~\cite{nips14:sutskever} for machine translation. A RNN encoder reads input sentence and updates its hidden state recurrently. Then a RNN decoder takes the last hidden state of the encoder to generate the target sentence. \cite{corr14:bahdanau} further enhances this architecture with an attention mechanism. Attention mechanism allows the decoder to generate each word by focusing on different parts of the input sentence. Hereafter, many neural dialogue sysetems have been developed. \cite{acl15:shang} devises a model to combine both the local and global information within the input sentence for response generation. Several works are proposed to exploit external knowledge for the purpose of more informative and relevant response generation~\cite{coling16:mou,aaai17:xing,emnlp18:pei}. \cite{aaai18:wu} proposes a dynaimc vocabulary mechanism to identify the possible response words for each input message. Very recently, \cite{icml17:gehring} proposes a convolutional network based Seq2Seq architecture. In their work, both encoder and decoder utilize stacked convolutional block structure to model a sentence. Unlike recurrent state update utilized by RNN, convolution operations do not require the hidden state at the previous steps and therefore enables parallelization over all words of a sentence. Because of this unique merit, we build the proposed \RAGE on the basis of this convolutional Seq2Seq architecture. 

The most relevant work to ours are the question answering systems presented in~\cite{ijcai16:yin,acl17:he,acl18:madotto,ijcai18:zhou}. \cite{ijcai16:yin} proposes a RNN based model to generate response for a simple factoid question. \cite{acl17:he} incorporates copying and retrieving mechanisms to accommodate multiple facts to generate a response for a complex question. \cite{acl18:madotto} combines multi-hop attention mechanisms with MemoryNN and incorporates KB information and dialog history into answer generation. Similarly, \cite{ijcai18:zhou} proposes a commonsense knowledge aware conversational model with static and dynamic graph attention. They encode structured and connected sementic information provided in the knowledge graph to facilitate response generation. However, these models need to retrieve the candidate factual information from a predefined knowledge base. This hinders its application in E-Commerce, because it is unknown how to model opinion information as (\textit{subject},~\textit{predict},~\textit{object}) triples. To the best of our knowledge, the propsoed \RAGE is the first work to generate a natural answer for a product-related question by exploting unstructured and noisy review data in E-Commerce. 
\section{Conclusion}
In this paper, we proposed a novel \textbf{r}eview-driven framework for \textbf{a}nswer \textbf{g}eneration for product-related questions in \textbf{E}-Commerce, named \RAGE. The experimental results over two real-world E-Commerce datasets demonstrate that \RAGE can identify the relevant information from the noisy review snippets and supervise the answer generation to produce more accurate and informative answers in natural language.

\begin{acks}
This research was supported by National Natural Science Foundation of China (No.61872278,No.61502344), Natural Science Foundation of Hubei Province (No.2017CFB502). Chenliang Li is the corresponding author.
\end{acks}

\bibliographystyle{ACM-Reference-Format}
\bibliography{ref}

\end{document}